\documentclass[letterpaper, 10 pt, conference]{ieeeconf}

\IEEEoverridecommandlockouts                              
\usepackage{amsmath}
\usepackage{amsfonts,amssymb}
\usepackage{bm}
\usepackage{graphicx}
\usepackage{subcaption}
\usepackage[linesnumbered,ruled]{algorithm2e}
\usepackage{multirow}
\usepackage{array}
\usepackage{color}
\usepackage{hyperref}
\usepackage[nameinlink]{cleveref}
\usepackage{textcomp}
\usepackage{balance}

\usepackage{bm}

\newcommand{\rem}[2]{{\textcolor{#1}{{#2}}}}

\newcommand{\jiexiong}[1]{\rem{black}{#1}}
\newcommand{\ludde}[1]{\rem{magenta}{#1}}

\newcommand{\sign}{\operatorname{sign}}

\title{\LARGE \bf GCNv2: Efficient Correspondence Prediction for Real-Time SLAM}
\author{Jiexiong Tang$^{1}$, Ludvig Ericson$^{1}$, John Folkesson$^{1}$ and Patric Jensfelt$^{1}$
\thanks{This work was partially supported by the Wallenberg AI, Autonomous Systems and Software Program (WASP), the SSF project FACT and the VR grant XPLORE3D.}
\thanks{$^{1}$The authors are all with the Centre for Autonomous Systems at KTH Royal Institute of Technology, Stockholm, SE-10044, Sweden
{\tt\small jiexiong@kth.se}}%
}

\begin{document}
 \thispagestyle{empty}
 \onecolumn
 \textcopyright 2019 IEEE.  Personal use of this material is permitted.  Permission from IEEE must be obtained for all other uses, in any current or future media, including reprinting/republishing this material for advertising or promotional purposes, creating new collective works, for resale or redistribution to servers or lists, or reuse of any copyrighted component of this work in other works.

\twocolumn
\newpage

	\maketitle

\begin{abstract}
In this paper, we present a deep learning-based network, GCNv2, for generation of keypoints and descriptors. GCNv2 is built on our previous method, GCN, a network trained for 3D projective geometry. GCNv2 is designed with a binary descriptor vector as the ORB feature so that it can easily replace ORB in systems such as ORB-SLAM2. GCNv2 significantly improves the computational efficiency over GCN that was only able to run on desktop hardware. We show how a modified version of ORB-SLAM2 using GCNv2 features runs on a Jetson TX2, an embedded low-power platform. Experimental results show that GCNv2 retains comparable accuracy as GCN and that it is robust enough to use for control of a flying drone. Source code is available at: \\ \protect\url{{https://github.com/jiexiong2016/GCNv2_SLAM}}
\end{abstract}

\section{Introduction}
The ability to estimate position is key to most, if not all, robotics application involving mobility. In this paper we focus on the problem of visual odometry (VO), i.e. 
relative motion estimation based on visual information. This is the corner stone in vision-based SLAM systems, like the one we demonstrate our method with. As in our previous work, \cite{GCN}, we estimate the motion using only an RGB-D sensor, and our target platform is a drone operating in an indoor environment. The RGB-D sensor makes scale directly observable without the need for visual-inertial fusion or the computational cost of inferring depth using a neural network as in \cite{Left-right,Single-stereo,SFMLearner}. This increases robustness, a key property for a drone. In particular, for indoor environments. Here the margin for error is small and there is typically less texture than outdoors. Our method is designed to be applicable to a system with an RGB-D sensor, without needing complicated calibration and synchronization with other sensors. Fusion can instead take place at a lower rate and with less need for precise timing, which, for example, makes integration with the flight control system of a drone simpler.

	\begin{figure}
	    \centering
	    \resizebox{\hsize}{!}{
		\includegraphics[width=0.5\textwidth]{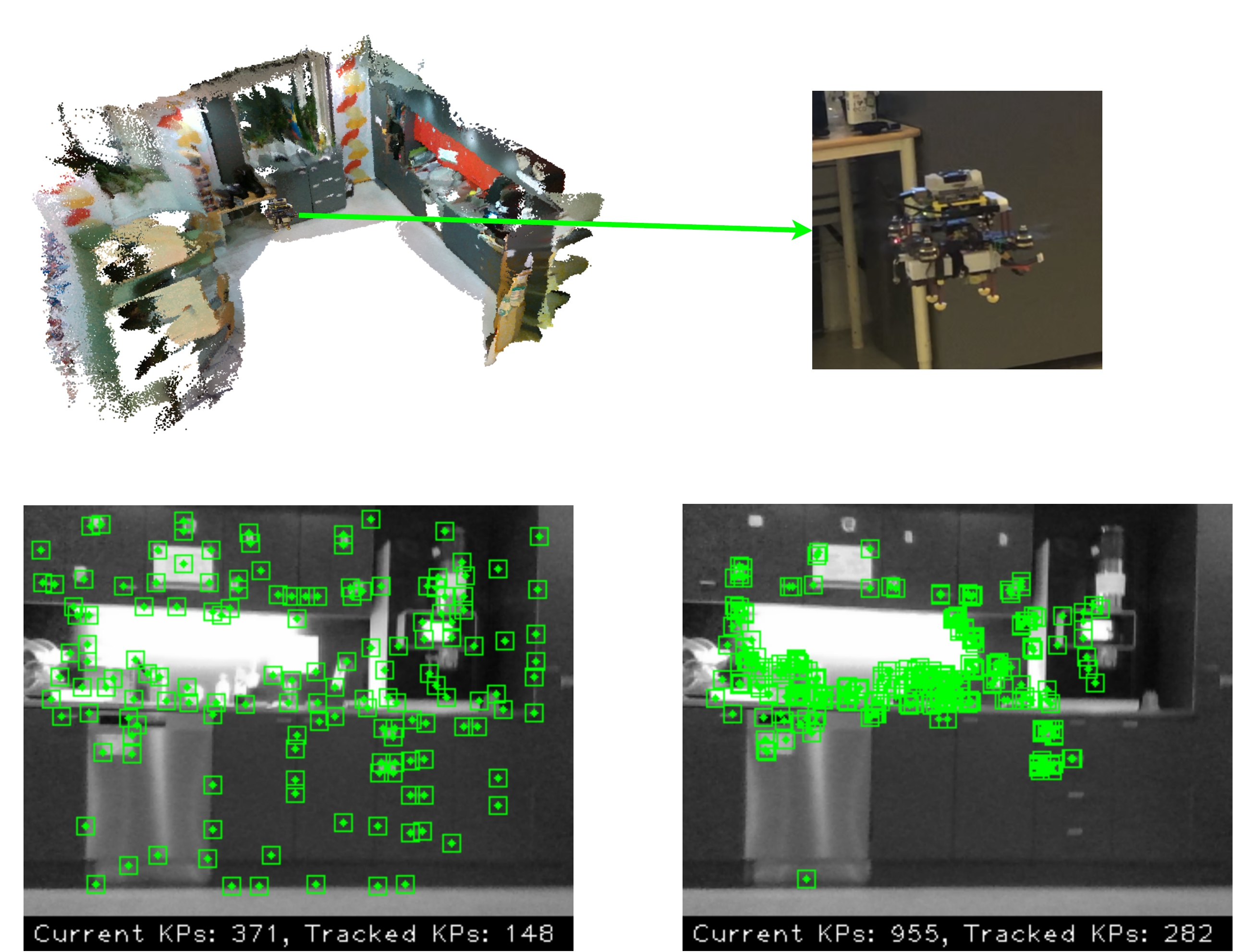}
		}
		\caption{The top figure shows our drone preforming position hold using GCN-SLAM. The figures below show the intermediate output for comparison of binary features, ORB and GCNv2, in ORB-SLAM2 and GCN-SLAM respectively. GCNv2 (left) tends to predict more repeatable and evenly distributed features compared with ORB (right.)}
		\label{fig:digest}
	\end{figure}

There is a trend in SLAM to investigate deep learning-based methods. 
In \cite{SuperPoint} a CNN based keypoint detector and descriptor called SuperPoint is presented. Experimental results show that SuperPoint has more distinctive descriptors than classical methods such as SIFT, and SuperPoint's detector is on par with classical methods. 
In our previous work~\cite{GCN} we introduced the Geometric Correspondence Network, GCN, specifically tailored for producing keypoints for camera motion estimation, achieving better accuracy than classical methods. However, due to the computational demands of GCN and its multi-frame matching setup, it is difficult to achieve real-time performance in a fully-operational SLAM system, e.g. on board a drone. Both keypoint extraction and matching are computationally too expensive. \jiexiong{Indeed, integration of deep learning into SLAM systems in performance-constrained settings is identified as an open problem in \cite{cadena2016past}.}%

In this paper we introduce GCNv2, based on the conclusions from~\cite{GCN}, to improve computational efficiency while still maintaining the high precision of GCN. \jiexiong{We rectify the multi-frame setup issue by instead predicting for a single frame at a time.} Our contributions are:
\begin{itemize}
\item \jiexiong{GCNv2 retains comparable accuracy to GCN, providing notable improvements in motion estimation compared to related deep learning-based feature extraction methods, while significantly reducing inference time.}
\item \jiexiong{We include binarization of the feature vector into the training, which greatly accelerates the matching. We design GCNv2 to have the same descriptor format as ORB features and have the ability to be used directly as the keypoint extractor in SLAM systems like ORB-SLAM2~\cite{ORBSLAM2-TRO17} or SVO2~\cite{SVO-TRO17}.}
\item We demonstrate the effectiveness and robustness of our work by using GCN-SLAM\footnote{Built on ORB-SLAM2 with ORB substituted for GCNv2} on a real drone for control, and show that it handles situations where ORB-SLAM2 fails. GCN-SLAM runs real-time on embedded low-power hardware, such as a Jetson TX2, as opposed to GCN which requires a desktop GPU for real-time inference.
\end{itemize}

\section{Related Work}
\label{sec:related}
In this section we cover related work in two areas. First VO and SLAM methods are covered, then we focus specifically on work on deep learning-based methods for image correspondence.

\subsection{VO and SLAM}
In direct methods for VO and SLAM, motion is estimated by aligning frames based directly on the pixel intensities, with~\cite{Comport07} being an early example. DVO (Direct Visual Odometry), presented in~\cite{DVO-Kerl13}, adds a pose graph to reduce the error. DSO~\cite{DSO} is a direct and sparse method that adds joint optimization of all model parameters. An alternative to the frame-to-frame matching is to match each new frame to a volumetric representation as in KinectFusion~\cite{KinectFusion}, Kintinous~\cite{Kintinous-IJRR14} and ElasticFusion~\cite{ElasticFusion}.

In indirect methods, the first step in a typical pipeline is to extract keypoints, which are then matched to previous frames to estimate the motion. The matching is based on the keypoint descriptors and geometric constraints. The state-of-the-art in this category is still defined by ORB-SLAM2~\cite{ORBSLAM-TRO15, ORBSLAM2-TRO17}. The ORB descriptor is a binary vector allowing high-performance matching. 

Somewhere between direct and indirect methods we find the semi-direct approaches. SVO2~\cite{SVO-TRO17} is a sparse method in this category, and can run at hundreds of Hertz. There are also semi-dense methods, in which category LSD-SLAM~\cite{LSD-SLAM} was one of the first. RGBDTAM~\cite{RGBDTAM} combines both semi-dense photometric and dense geometric errors for pose estimation.

There are a number of recent deep learning-based mapping systems like \cite{CNN-SLAM, DVSO}. The focus in these methods is deep learning-based single view depth estimation to reduce the scale drift inherent in monocular systems. CNN-SLAM~\cite{CNN-SLAM} feeds the depth into LSD-SLAM. In DVSO~\cite{DeepVO}, depth is predicted in a similar way to \cite{Left-right}, using a virtual stereo view. CodeSLAM \cite{code-slam} learns an optimizable representation from conditioned auto-encoding for 3D reconstruction. In S2D~\cite{S2D}, we build on DSO~\cite{DSO} and exploit both depth and normals predicted by a jointly optimized CNN. Some work on unsupervised training for motion estimation also exist. Image reconstruction loss is used for unsupervised learning in~\cite{SFMLearner, Undeepvo}. However, geometry-based optimization methods still outperform end-to-end systems as shown in~\cite{DeepVO}.

\subsection{Deep Correspondence Matching}
There is a an abundance of recent works that deploy variants of metric learning for training deep features for finding image correspondences~\cite{feat_ref1, feat_ref2, feat_ref3, feat_ref4, feat_ref5, feat_ref6, LIFT, UCN, SuperPoint}.
Works in~\cite{cd_ref1, cd_ref2} focus on improving learning-based detection with better invariances. 
Aimed at a different aspect, \cite{cd_ref3, cd_ref4, cd_ref5} use synthetic samples generated in a self-supervised manner to improve general feature matching.

Among the aforementioned methods, LIFT~\cite{LIFT} in particular uses a patch-based method to perform both keypoint detection and descriptor extraction. SuperPoint~\cite{SuperPoint} predicts the keypoints and descriptors using a single network together with the self-supervised strategy in~\cite{cd_ref5}. Notably, \cite{SuperPoint} shows that \cite{SuperPoint, LIFT, UCN} work on par with classical methods like SIFT for motion estimation.

In GCN~\cite{GCN}, we show that by learning keypoints and descriptors specifically targeting motion estimation, performance is improved -- contrary to what is reported for other more general deep learning-based keypoint extractor systems~\cite{SuperPoint,UCN}. In this paper we introduce an high-throughput variant to GCN, dubbed GCNv2. We demonstrate the applicability of these keypoints for SLAM and build on ORB-SLAM2 as it offers a comprehensive multi-threaded state-of-the-art indirect SLAM system with support for monocular as well as RGB-D cameras. ORB-SLAM2 complements the tracking front-end with a back-end that does both pose graph optimization using g2o~\cite{g2o} and loop closure detection using a binary bag of words representation~\cite{Galves-BoW-TRO12}. To simplify this integration, we design the GCNv2 descriptor to have the same format as that of ORB.

\section{Geometric Correspondence Network}

In this section we present the design of GCNv2, aimed at making GCN suitable for real-time SLAM applications running on embedded hardware. We first introduce the revised network structure, then detail the training scheme used for the binarized feature descriptor and keypoint detector.


\subsection{Network Structure}

\begin{figure}[!t]
    \centering
    \resizebox{1.0\hsize}{!}{
        \centering
        \includegraphics{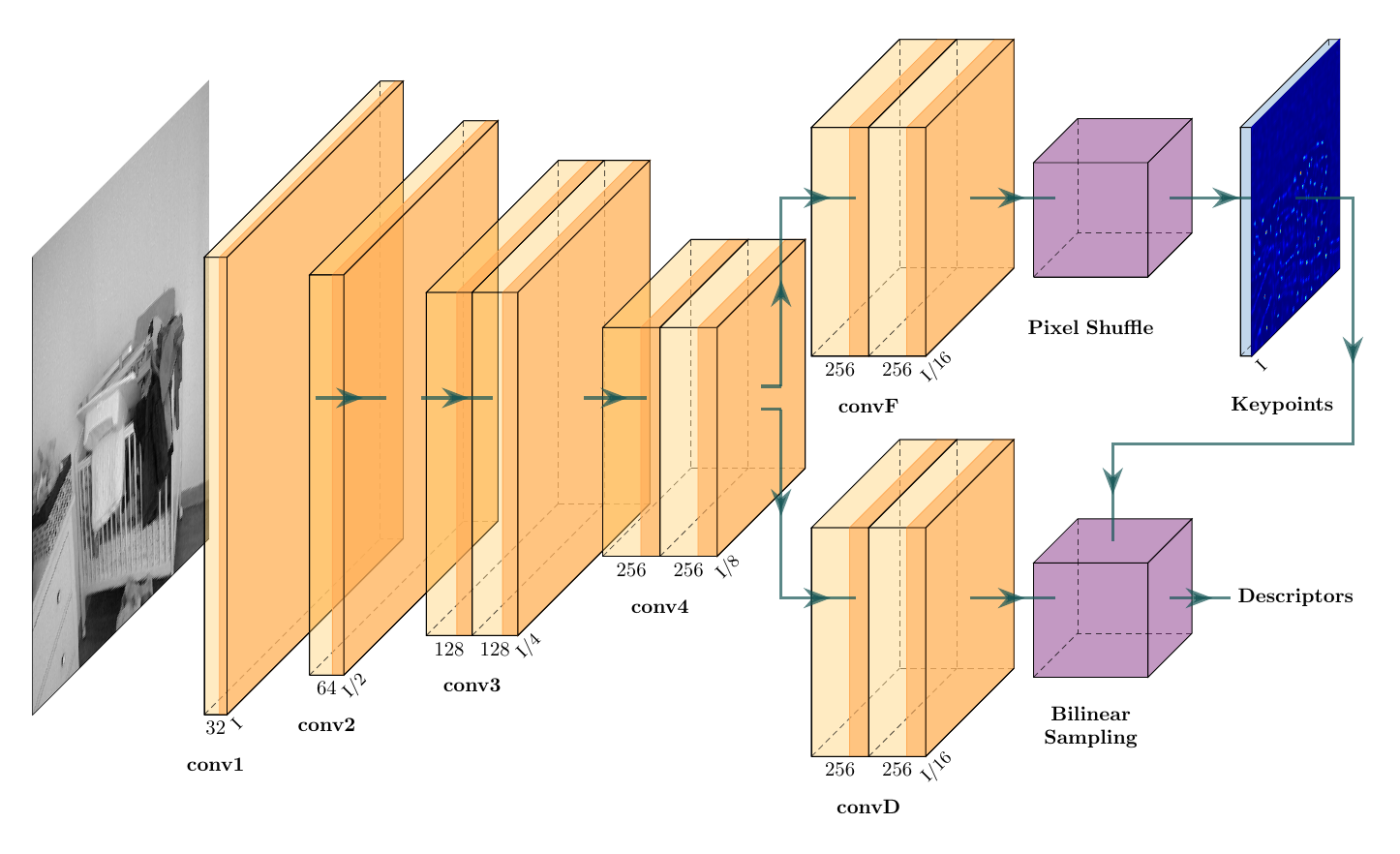}
    }
    \caption{\jiexiong{The GCNv2 network structure. The orange blocks represent convolutional layers, and numbers underneath are corresponding number of channels. The ELU non-linearity is applied after each set of blocks. The notation I divided by some number indicates the resolution of the current feature map, e.g. I/4 means one pixel is 4x4 pixels in the input image.}}
    \label{fig:structure}
\end{figure}

The original GCN structure, proposed in~\cite{GCN}, consists of two main parts: an FCN~\cite{FCN} structure with a ResNet-50 backbone, and a bidirectional recurrent convolutional network. The FCN is adapted for dense feature extraction, whereas the bidirectional recurrent network is used to find keypoint locations. \cite{GCN} shows that GCN has impressive tracking performance compared to existing methods, however it also notes that the formulation has practical limitations as applied to real-time SLAM systems. We identify two main issues: first, the network architecture is fairly large, and so requires powerful computational hardware which renders it unable to run in real-time on board e.g. the Jetson TX2 used for our drone experiments; and second, GCN inference requires two or more frames be input at the same time, increasing not only computational but also algorithmic complexity.

\jiexiong{To address these limitations we introduce a single view based simplified network structure, GCNv2, with improved efficiency. 
The overall GCNv2 network structure is shown in \cref{fig:structure}. 
Like GCN, GCNv2 predicts both keypoints and descriptors at the same time, i.e. the network outputs a probability map for keypoint confidences and a dense feature map for descriptors.
Inspired by SuperPoint~\cite{SuperPoint}, GCNv2 is modified to perform predictions at a low resolution rather than the original one, and only uses a single image. This is achieved by first predicting the probability map and dense feature map at low resolution, then pixel shuffling the $256$ channels probability map\footnote{The resolution at \textit{convF} in \cref{fig:structure} is $20 \times 15$, the number of channels for \textit{convF} must be $16\times 16$ to recover the original resolution, i.e., $320\times240$.} to the original resolution, and finally performing non-maximum suppression over the full resolution probability map and using these locations to bilinearly sample the corresponding feature vectors from the low resolution dense feature map (right part in \cref{fig:structure}.) }

\jiexiong{GCNv2 together with our adapted SLAM method, GCN-SLAM, runs at around 80 Hz on a laptop with Intel i7-7700HQ and mobile version NVIDIA 1070. To achieve an even higher frame rate necessary for real-time inference on the Jetson TX2, we introduce a smaller version of GCNv2, called GCNv2-tiny, where we reduce the number of feature maps by half from \textit{conv2} and onward. GCNv2-tiny runs at 40 Hz and the GCN-SLAM using it runs 20 Hz on the TX2, is therefore well-suited for deployment on a drone. In addition to the mobile platforms, a more comprehensive comparison of the inference and matching time on a desktop computer is shown in~\cref{fig:efficiency}. It can be seen that the resolution of the input affects the inference time in a coarsely quadratic manner. GCNv2 achieves a lower inference time compared to GCN and SuperPoint at the same resolution. This is mainly due to our modifications to the network architecture. Further details on the GCNv2 and GCNv2-tiny network specifics can be found in our publicly available source code\footnote{\url{https://github.com/jiexiong2016/GCNv2_SLAM/blob/master/GCN2/Network.md}}.
}

\subsection{Feature Extractor}

\begin{figure}[!t]
\includegraphics[width=0.5\textwidth]{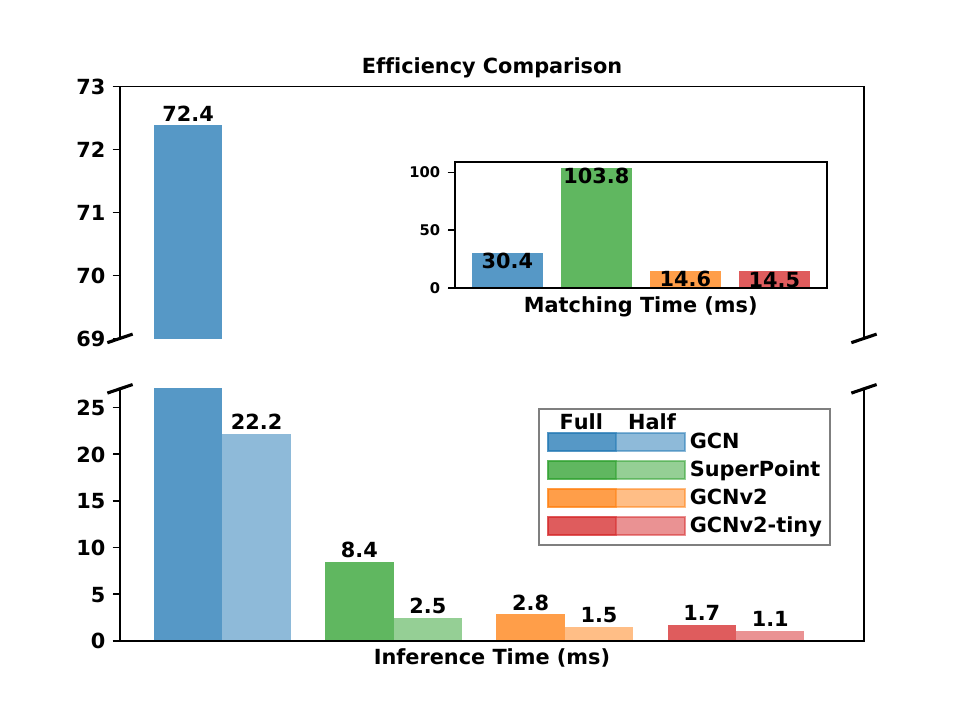}
\caption{\jiexiong{Efficiency comparison of different methods. For a fair comparison, the matching time is evaluated by performing nearest neighbour matching using the same number (1000) of keypoints for all methods. The dark and light colors indicate inference time of input with full resolution ($640\times480$) and half resolution ($320\times256$ for GCN since it is based on ResNet and $320\times240$ for the rest.)}}
\label{fig:efficiency}
\end{figure}

\textbf{\jiexiong{Notation}} \jiexiong{We firstly introduce some important notation which will be used in the following subsections. We denote $\bm{f}$ as the dense feature map at the low resolution and $\bm{o}$ as the pixel shuffled keypoint probability map, i.e., at original resolution. When giving a location $\bm{x} = (u, v)$, 2D coordinates in image plane, as input to $\bm{f}(\cdot)$ and $\bm{o}(\cdot)$, corresponding feature descriptor and keypoints confidence are returned. Specifically, the descriptor is obtained using bilinear sampling. Superscript indicates the input frame, e.g., $\bm{f}^{cur}$ and $\bm{f}^{tar}$ represent feature map of current and target frames, respectively.
Finally, $L_{feat}$ and $L_{det}$ are losses for training the descriptor and detector, respectively.}

\textbf{Binarized Descriptor}
We trained the features of GCNv2 to be binary for accelerating the matching procedure and to match those of ORB. To binarize the features, we add a binary activation layer on top of the final output. It is essentially a hard sign function non-linearity and is therefore not differentiable everywhere. The challenge is how to back-propagate the loss properly through this layer of the network. We used the method proposed in~\cite{Binarization}. The binary activation layer can be written as follows:
\begin{equation} \label{eq:bin_activation}
\begin{split}
\text{Forward: }\,\bm{b}(\bm{x}) &= \sign(\bm{f}(\bm{x})) = 
\left\{
\begin{array}{c l}	
     +1 & \bm{f}(\bm{x}) \ge 0\\
     -1 & \text{otherwise}
\end{array}\right. \\
\text{Backward: }\,\frac{\partial \bm{b}}{\partial \bm{f}} &=   \bm{1}_{|\bm{f}| \le 1}
\end{split}
\end{equation}
\jiexiong{where $\bm{b}$ is the binarized version of feature $\bm{f}$.  $\bm{1}_{|\bm{f}| \le 1}$ cancels the gradients for individual feature responses of $\bm{f}$ with absolute values greater than $1$. It is a so-called \textit{straight-through estimator} of the hard sign function for backpropagating the gradients~\cite{Binarization}. This indicator function approximates binarization during backprogation and avoids over-penalization to feature responses already passed the decision boundary by $1$.} We found that it is more efficient to train the network with the above method rather than forcing the network to directly predict a binary output by minimizing quantification loss as in \cite{Lin-DeepBit-CVPR16}. One possible reason is that forcing the value to be clustered around $
\{+1,-1\}$ conflicts with the metric learning that follows.

\jiexiong{The size of the binary feature vector $\bm{b}$ is set to $256$, the same as ORB, making it trivial to directly incorporate GCNv2 into existing ORB-based visual tracking systems such as ORB-SLAM2. As shown in \cref{fig:efficiency}, the matching speed with binarized features far outperforms that of SuperPoint for the same feature vector size and GCN with 64 in feature vector size.}


\textbf{Nested Metric Learning} Pixel-wise metric learning is used to train the descriptor in a nearest-neighbour manner. The triplet loss for binarized features is as follows:
\begin{equation} \label{eq:triplet}
\begin{gathered}
L_{feat} = \sum_{i} \max (0, d(\bm{x}^{cur}_i, \bm{x}^{tar}_{i,+}) - d(\bm{x}^{cur}_i, \bm{x}^{tar}_{i,-}) + m) \\
d({\bm{x}^{cur}, \bm{x}^{tar}}) = || \bm{b}^{cur} ( {\bm{x}^{cur}} )  - \bm{b}^{tar} ( {\bm{x}^{tar}} ) ||_2
\end{gathered}
\end{equation}
\jiexiong{where $m$ is the distance margin for truncation. $d(\cdot,\cdot)$ is the squared Hamming distance for the binarized features.} We use squared distance as we found that it results in faster and better convergence for training. $(\bm{x}^{cur}_i, \bm{x}^{tar}_{i,+})$ is a matching pair obtained using the ground truth camera poses from the training data as follows:
\begin{equation} \label{eq:cor_gt}
\bm{x}^{tar}_{i,+} =  \bm{\pi}^{-1}(\mathbf{R}_{gt} \bm{\pi}(\bm{x}^{cur}_i, z_i) + \bm{t}_{gt})
\end{equation}
\jiexiong{where $\mathbf{R}_{gt} \in \mathbb{R}^{3 \times 3}$ is the ground truth rotation matrix and $\bm{t}_{gt} \in \mathbb{R}^3$ is the ground truth translation vector. $\bm{\pi}$ unproject a pixel from the image plane to 3D space using the given 2D coordinates $\bm{x}^{cur}_i$ and depth $z_i$. 
$(\bm{x}^{cur}_i, \bm{x}^{tar}_{i,-})$ is a non-matching pair retrieved by exhaustive negative sample mining described in \cref{alg:ENSM}}. The exhaustive search will further penalize the already matched features with the relaxed criteria described in~\cite{GCN}. The relaxed criteria is used to increase the tolerance to potentially noisy data.

\addtolength{\topmargin}{2mm}
\begin{algorithm}[!t]
\caption{Exhaustive Negative Sample Mining}
\label{alg:ENSM}
\SetKwInOut{Input}{Input}
\SetKwInOut{Output}{Output}
\Input{Current $i$th feature: $\bm{b}^{cur} ( \bm{x}^{cur}_{i} )$, \\
        $k$ nearest features: $ \{\bm{b}^{tar} ( \bm{x}^{tar}_{j} ) \, | \, j\in[1,k] \} $, \\ 
       Ground truth correspondence: $\bm{x}^{tar}_{i, +}$, \\
       Relaxed criteria: $c$}
\Output{Negative sample $\bm{b}^{tar}( \bm{x}^{tar}_{i,-} ) $ \textbf{or} None}
\For{$n=1;n\leq k;n\text{\scshape++}$}{
    \If{${||\bm{x}^{tar}_{n} - \bm{x}^{tar}_{i,+}||}_1 > c $}{
        \Return $\bm{b}^{tar} ( \bm{x}_{n}^{tar} )$
    }
}
\Return None
\end{algorithm}
 
\subsection{Distributed Keypoint Detector}
\jiexiong{Similar to GCN, we treat the keypoint detection as a binary classification problem. The target of the probability map $\bm{o}$ from the network is a mask consisting of $1$ and $0$. These values indicate whether a pixel is a keypoint or not. Weighted cross-entropy is then used as objective function for the training. The loss is always evaluated on two consecutive frames, with the aim of enhancing the consistency of extracted keypoints. The loss for the detection can be written as follows:}	
\begin{equation} \label{eq:NLL}
\begin{gathered}
	L_{det} = L_{ce}(\bm{o}^{cur}( \bm{x}^{cur})) + L_{ce}(\bm{o}^{tar}( \bm{x}^{tar}_{+})) \\
	\begin{split}
	L_{ce}(\bm{o}(\bm{x})) = -& \alpha_1 \sum_{i} c_{\bm{x}_i} \mathrm{log}(\bm{o}({\bm{x}_i})) \\ 
	                         -& \alpha_2 \sum_{i} (1-c_{\bm{x}_i}) \mathrm{log}(1-\bm{o}({\bm{x}_i}))
	\end{split}
\end{gathered}
\end{equation}	
\jiexiong{where $\alpha_1$ and $\alpha_2$ are used for handling the unbalanced classes to prevent pixels which are not keypoints from dominating the loss.} We generate the ground truth by detecting Shi-Tomasi corners in a $16\times{}16$ grid and warp them to the next frame using \cref{eq:cor_gt}.  This leads to better distribution of keypoints and the objective function directly reflects the ability to track the keypoints based on texture.

\subsection{Training Details}

\jiexiong{The final loss for the training is a weighted combination of $L_{feat}$ and $L_{det}$, with weights ${100}$ and $1$, respectively. These weights are selected to balance the scale of the two terms. The triplet loss margin $m$ is set to $1$. Relaxed criteria $c$ for exhaustive negative sample mining is set to $8$. The cross-entropy weights $[\alpha_1, \alpha_2]$ are set to $[0.1, 1.0]$.
The learning rate for the adaptive gradient descent method, ADAM~\cite{ADAM}, is started from $10^{-4}$ and halved every 40 epoch for a total of 100 training epochs. The weights of GCNv2 are randomly initialized under uniform distribution. }

\section{GCN-SLAM}
\label{sec:GCN-SLAM}

One of the most important design choices for a keypoint-based SLAM system is the choice of keypoint extractor. The keypoints are often re-used at multiple stages in such systems. ORB features~\cite{ORB}, the namesake of ORB-SLAM2, are a robust method as they are cheap to compute compared to other keypoint detectors with equivalent properties, and have a compact descriptor for fast matching. 

As previously shown in~\cite{GCN}, GCN used in a naive motion estimation pipeline performs better than or on par with ORB-SLAM2~\cite{ORBSLAM2-TRO17}. Notably, this is without higher-order SLAM functionality such as pose graph optimization, global bundle adjustment, or loop detection. Incorporating GCN into a system with such functionality would therefore be likely to yield better results. However, as mentioned, GCN is prohibitively expensive for  our target of real-time use on embedded hardware. In what follows we show how we modify ORB-SLAM2 to incorporate GCNv2, in a system we call GCN-SLAM.

\begin{figure}[!t]
    \centering
    \resizebox{0.7\columnwidth}{!}{
    \includegraphics{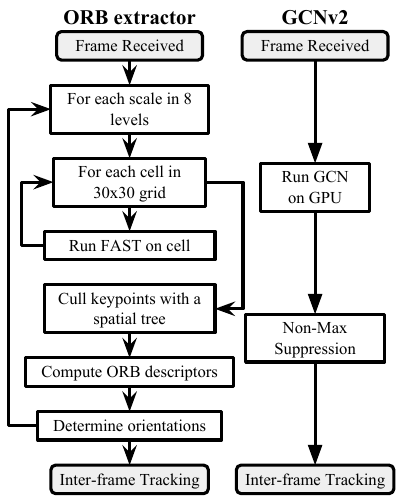}
    }
    \caption{Illustration of the original ORB-SLAM2 keypoint extraction process on the left side, and our method on the right. The keypoint extraction in GCN-SLAM is comparatively simple, in large part because it relies on 2D convolutions and matrix multiplication which is off-loaded to the GPU.}
    \label{fig:piplines}
\end{figure}



ORB-SLAM2's motion estimation is based on frame-to-frame keypoint tracking and feature-based bundle adjustment. We will briefly describe the detection and description of these features. ORB-SLAM2 employs a scale pyramid where the input image is iteratively scaled down to enable multi-scale feature detection by running single-scale algorithms on the multiple rescaled images. For each scale level, the FAST corner detector is applied in a $30\times{}30$ grid. If no detections are found in a cell, FAST is run again with a decreased threshold. After all detections have been gathered from all cells at a given level in the scale pyramid, a space partitioning algorithm is used to cull the keypoints first by their image coordinates, then by detection score. Finally, once typically 1000 keypoints have been selected in total, the viewing angle of each keypoint is computed, then each pyramid scale level is filtered with Gaussian blur, and the 256-bit ORB descriptor for each keypoint at each level is computed from the blurred image.

Our method computes both keypoint locations and descriptors simultaneously in a single forward-pass of the network, and as stated before, its end result is designed to be a drop-in replacement for the ORB feature extractor outlined above. 
The two keypoint methods are illustrated in \cref{fig:piplines}.

Once keypoints and their respective descriptors are found, ORB-SLAM2 relies primarily on two methods for frame-to-frame tracking: first, by assuming constant velocity and projecting the previous frame's keypoints into the current frame, and if that fails, by matching the keypoints of the current frame to the last-created keyframe using bag-of-words similarity. We have disabled the former so as to use only the latter keypoint-based reference frame tracking. We have also replaced the matching algorithm with a standard nearest-neighbor search in our experiments. These modifications are made to examine the performance of our keypoint extraction method, rather than that of ORB-SLAM2's other tracking heuristics. 

Finally, we have left ORB-SLAM2's loop closure and pose graph optimization intact, apart from having regenerated the bag-of-words vocabulary to suit GCNv2 feature descriptors by computing them on the training dataset presented in \cref{sec:trainingdata}.


\section{Experimental Results}

\addtolength{\topmargin}{2mm}
\begin{table*}[!ht]
\caption{ATE USING FRAME TO FRAME TRACKING}\label{tab:ate_open}
\centering
\begin{tabular}{ l | c | c | c | c || c | c | c | c}
\hline
Dataset (200 Frames) & GCN & ORB & SIFT & SURF & SuperPoint & GCNv2 & GCNv2-tiny & GCNv2-large\\ 
\hline \hline
fr1\textunderscore floor & \textbf{0.015m} & 0.080m  &  0.073m  & 0.074m & -  & -  & -& -\\  \hline
fr1\textunderscore desk & \textbf{0.037m} & 0.151m  & 0.144m & 0.148m & 0.166m & 0.049m & 0.084m & 0.038m \\  \hline
fr1\textunderscore 360 & \textbf{0.059m} & 0.278m & 0.305m  & 0.279m  & - & - & - & 0.097m \\  \hline
fr3\textunderscore long\textunderscore office& 0.061m & 0.090m  &  0.076m & 0.070m & 0.105m & \textbf{0.046m} & 0.085m & 0.067m\\ \hline
fr3\textunderscore large\textunderscore cabinet & 0.073m & 0.097m &  0.091m & 0.143m & 0.195m & 0.064m & 0.067m & \textbf{0.056m} \\ \hline
fr3\textunderscore nst & 0.020m &  0.061m &  0.036m & 0.030m & 0.055m & \textbf{0.018m} & 0.024m & 0.021m\\ \hline
fr3\textunderscore nnf & \textbf{0.221m} & -  &  - & - & - & - & - & -\\ \hline
\end{tabular}
\end{table*}

In this section, we present experimental results to justify our conclusions regarding the performance of our keypoint extraction method, and its embodiment in the GCN-SLAM system. 
\jiexiong{Our work is not intended as an alternative to ORB-SLAM2, but rather a keypoint extraction method that is: i) tailored to motion estimation, ii) computationally efficient, and iii) suitable for use in a SLAM system.
\Cref{sec:quantitative} presents our quantitative results by benchmarking GCN-SLAM against ORB-SLAM2 and some related methods. \Cref{sec:qualitative} we qualitatively compare our method to ORB features by using them in the same SLAM pipeline as described in \cref{sec:GCN-SLAM}.}

For quantitative experiments, the evaluation is performed using a laptop with an Intel i7-7700HQ processor and a mobile version of NVIDIA 1070. For qualitative experiments, real-world scenarios, we used an NVIDIA Jetson TX2 embedded computer for processing and an Intel RealSense D435 RGB-D camera sensor on a custom-built drone (see \cref{fig:digest}.)


\subsection{Training Data}
\label{sec:trainingdata}
The original GCN was trained using the TUM dataset~\cite{TUM} from sensor fr2. It provides accurate camera poses through a motion capturing system. In GCNv2, we trained the network using a subset of the SUN-3D~\cite{SUN-3D} dataset we created in our recent work~\cite{S2D}. SUN-3D contains millions of real-world recorded RGB-D images in various typical indoor environments. A total of $44,624$ frames were extracted by roughly one frame per second. 
SUN-3D is a very rich dataset, and its diversity could potentially produce a more generalized network. However, the ground truth poses provided are estimated by visual tracking with loop closure and so are relatively accurate in a global sense, but have misalignments at the frame level. To account for this local error, we extract SIFT features and use the provided poses as initial guesses for bundle adjustment to update the relative pose of each frame pair. In this sense, the training of GCNv2 is using self-annotated data from the RGB-D camera. 

\subsection{Quantitative Results}
\label{sec:quantitative}
For comparison with the original GCN, we select the same sequences of the TUM datasets as in~\cite{GCN} and evaluate tracking performance with an open and a closed loop system. We use the Absolute Trajectory Error~(ATE)~\cite{TUM} as the metric.
Since we trained GCNv2 on a different dataset than the original GCN~\cite{GCN}, we also show results using the original recurrent structure for comparison. We have therefore also created GCNv2-large, with ResNet-18 as the backbone and deconvolutional up-sampling for the feature maps. The bidirectional feature detector is moved to the lowest scale as the other two versions of GCNv2.

\begin{table}[!ht]
\centering
\caption{ATE USING CLOSED LOOP SYSTEM}\label{tab:ate_closed}
\resizebox{\hsize}{!}{\begin{tabular}{ l | c | c | c | c || c } 
\hline
\multirow{2}{*}{Dataset} & \multirow{2}{*}{GCN} & ORB       & Elastic        & RGBD    & GCN     \\
                         &                      &     SLAM2 &         Fusion &     TAM &     SLAM\\
\hline \hline
fr1\textunderscore floor & 0.038m & 0.036m  &  - &  - & \textbf{0.021m} \\ \hline
fr1\textunderscore desk & 0.029m & \textbf{0.016m}  & 0.020m & 0.027m & 0.031m \\ \hline
fr1\textunderscore 360 & \textbf{0.069m} & 0.213m & 0.108m  & 0.101m & 0.155m \\ \hline
fr3\textunderscore long\textunderscore office & 0.040m & \textbf{0.010m}  &  0.017m & 0.027m & 0.021m \\ \hline
fr3\textunderscore large\textunderscore cabinet & 0.097m &  - &  0.099m & \textbf{0.070m} & \textbf{0.070m} \\ \hline
fr3\textunderscore nst & 0.020m &  0.019m &  0.016m & \textbf{0.010m} & 0.014m\\ \hline
fr3\textunderscore nnf & \textbf{0.064m} & -  &  -   &  - & 0.086m \\ \hline
\end{tabular}}
\end{table}

Frame-to-frame tracking results are shown in \cref{tab:ate_open}.
 \jiexiong{The columns left of the double vertical lines are from~\cite{GCN} where $640\times480$ images were used, the columns right of the double vertical lines are with image resolution halved, i.e. $320\times240$, since this is the resolution we used on our drone.}
\jiexiong{The results are consistent with those reported in \cite{SuperPoint}. SuperPoint performs on par with classical methods like SIFT, while GCNv2 performs close to GCN, but notably better than SuperPoint.}
GCNv2 performance is on par with GCN, and even slightly better in two cases, likely due to using a larger training dataset as described in \cref{sec:trainingdata}. The exceptions are fr1\textunderscore floor and fr1\textunderscore 360. These sequences require fine details, and as GCNv2 performs detection and descriptor extraction with lower scale feature maps, performance suffers accordingly. Further corroborating this is the fact that GCNv2-large successfully tracks fr1\textunderscore 360. Finally, we note that the smaller version of GCNv2, GCNv2-tiny, is only slightly less accurate than GCNv2. 

In \cref{tab:ate_closed}, we compare the closed loop performance of GCN-SLAM with our previous work, as well as ORB-SLAM2, ElasticFusion, and RGBD-TAM. GCN-SLAM succeeds in tracking for all sequences with an error comparable to that of GCN, whereas ORB-SLAM2 fails on two sequences. GCNv2 has less error than ORB-SLAM2 in the fast rotations of fr1\textunderscore360. It is also noteworthy that for this particular sequence, the original GCN does significantly better than both ORB-SLAM2 and GCN-SLAM. ORB-SLAM2 is tracking well in all other sequences, and the errors of both GCN-SLAM and ORB-SLAM2 are small. \jiexiong{These results are particularly encouraging, as GCN-SLAM does not use the ORB-specific feature matching heuristic that exist in ORB-SLAM2, leaving room for further performance improvements.}

\subsection{Qualitative Results}
\label{sec:qualitative}

\begin{figure*}[ht]
\centering
\begin{subfigure}{0.24\textwidth}
\centering
\includegraphics[width=\textwidth]{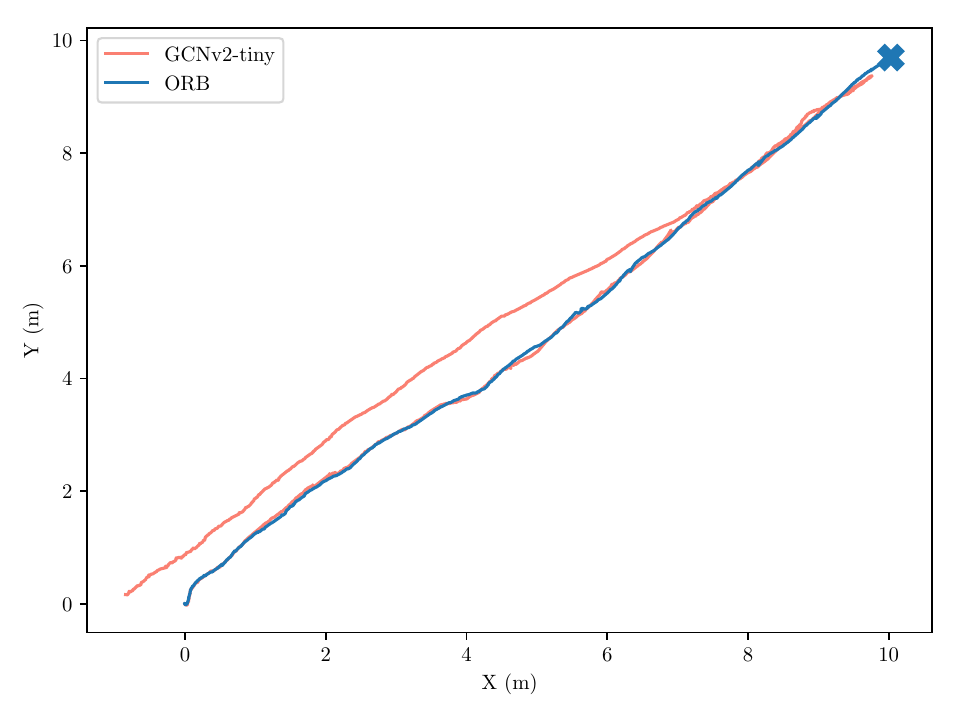}
\caption{\textit{Corridor}: indoor, handheld camera.}
\label{fig:trajs_cor}
\end{subfigure}
\hfill
\begin{subfigure}{0.24\textwidth}
\centering
\includegraphics[width=\textwidth]{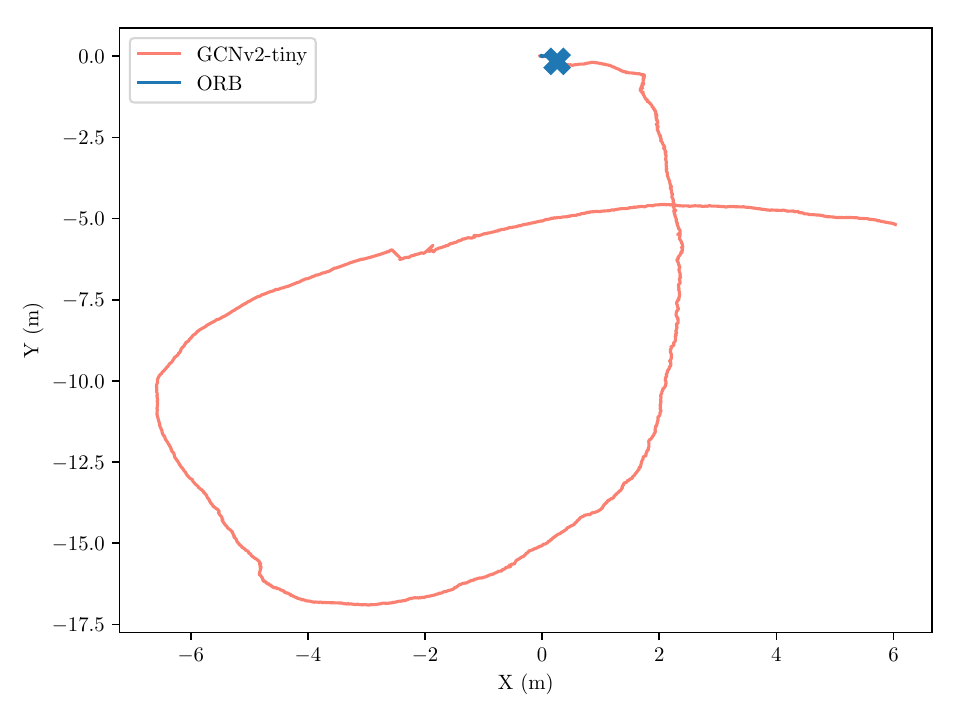}
\caption{\textit{Parking lot}: outdoor, handheld camera.}
\label{fig:trajs_park}
\end{subfigure}
\hfill 
\begin{subfigure}{0.25\textwidth}
\centering
\includegraphics[width=\textwidth]{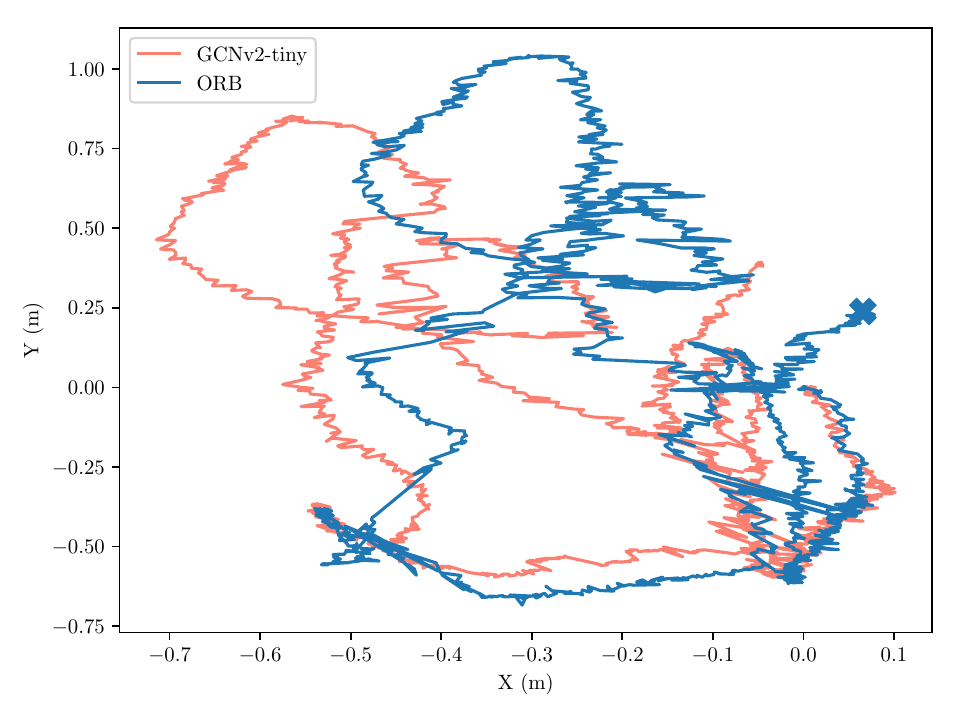}
\caption{\textit{Alcove}: indoor, flying with a primitive optical flow sensor.}
\label{fig:trajs_atrium}
\end{subfigure}
\hfill
\begin{subfigure}{0.24\textwidth}
\centering
\includegraphics[width=\textwidth]{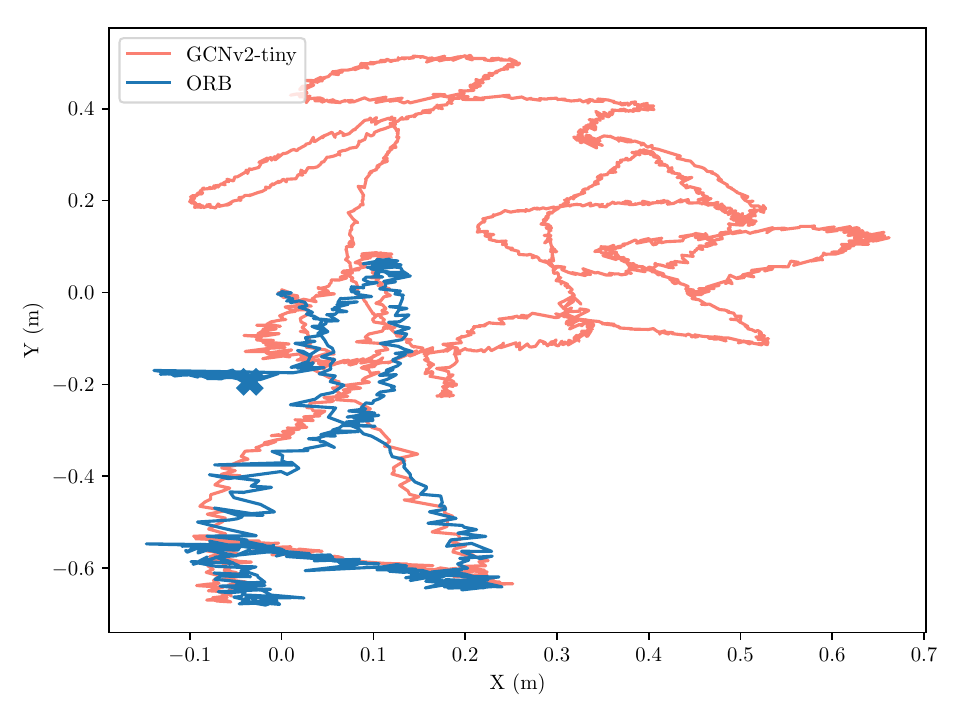}
\caption{\textit{Kitchen}: indoor, flying with GCN-SLAM for positioning.}
\label{fig:trajs_kitchen}
\end{subfigure}
\caption[Qualitative results]{Estimated trajectory on each of the four datasets using the GCN-SLAM pipeline and illustrating the difference between GCNv2 and ORB features. Note that there is no ground truth for these trajectories but we mark tracking failure with a cross and we see that this happened in all four cases when using ORB features. A demonstration video of our system is available online\footnotemark.}
\label{fig:trajs}
\end{figure*}
\footnotetext{\url{https://www.youtube.com/watch?v=pz-gdnR9tAM}}

\begin{figure*}[ht]
\centering
\begin{subfigure}{0.45\textwidth}
\centering
\includegraphics[width=\textwidth]{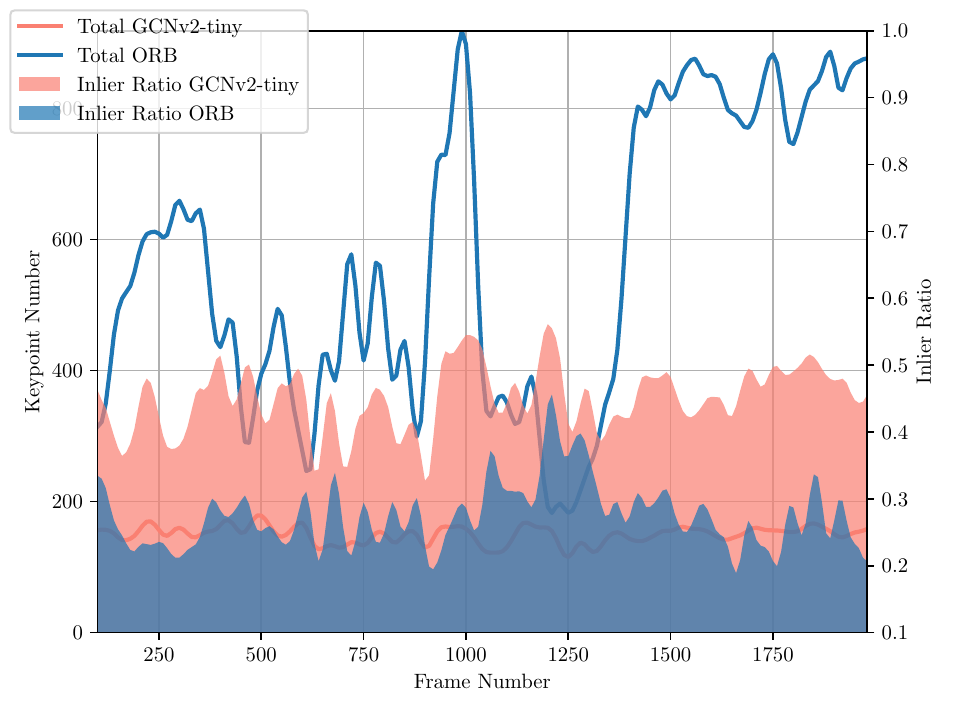}
\end{subfigure}
\begin{subfigure}{0.45\textwidth}
\centering
\includegraphics[width=\textwidth]{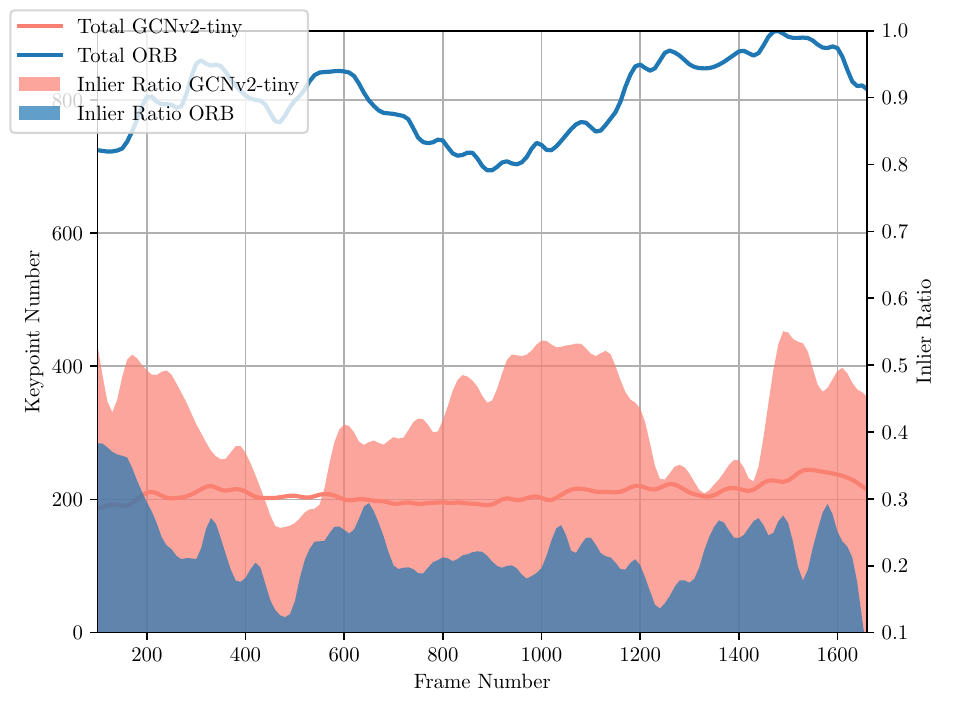}
\end{subfigure}
\caption{Keypoint extractor performance in terms of tracking shown for the \textit{Corridor} (left) and \textit{Kitchen} (right) datasets. Lines indicate the total number of keypoints detected per frame. The filled-in areas indicate the fraction of keypoints that were successfully used for local map tracking, i.e. contribute to tracking and is plotted against the right-hand axis.}
\label{fig:inliers}
\end{figure*}

To further verify the robustness of using GCNv2 in a real-world SLAM setting, we show results on four datasets collected in our environment under different conditions: a) going up a corridor, turning 180 degrees and walking back with a handheld camera, b) walking in a circle on an outdoor parking lot with a handheld sensor in daylight, c) flying in an alcove with windows and turning 180 degrees, and d) flying in a kitchen and turning 360 degrees while using GCN-SLAM for positioning. 

 \begin{figure*}[ht]
 \centering
 \begin{subfigure}[c]{0.49\textwidth}
 \centering
 \includegraphics[width=\textwidth]{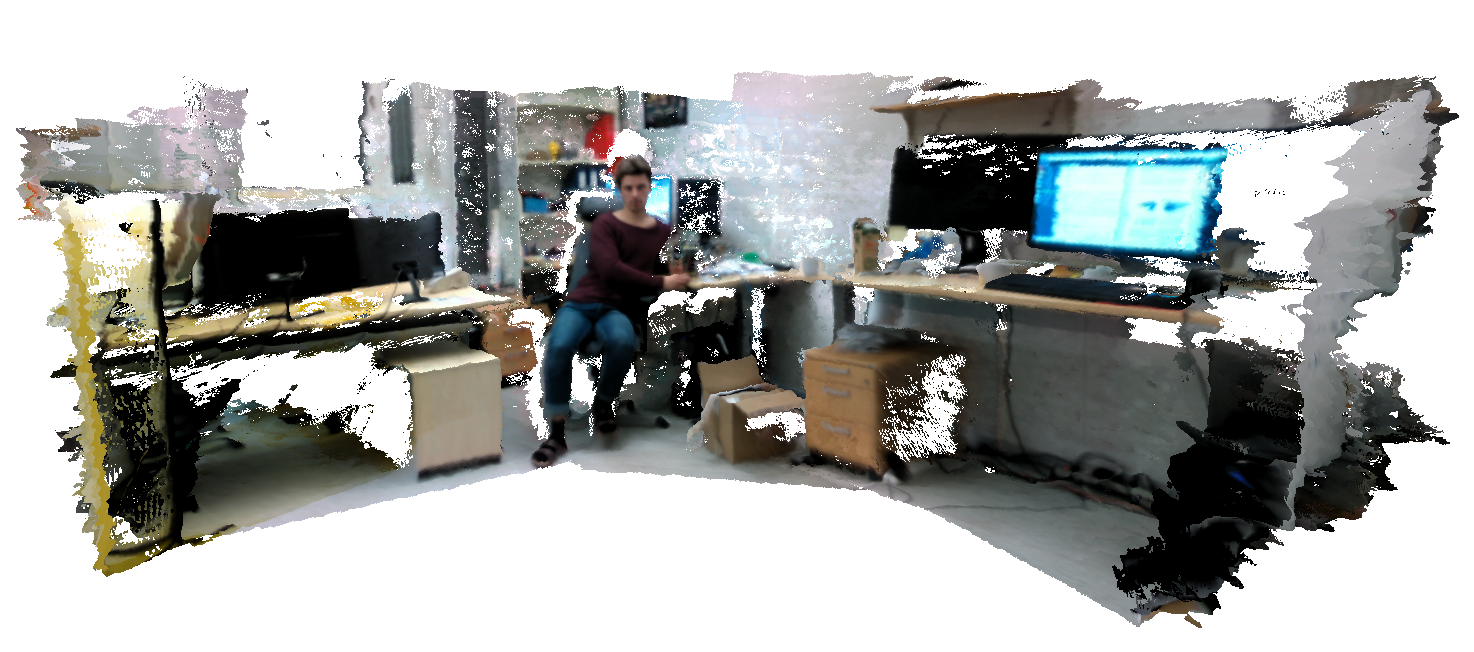}
 \end{subfigure}
 \centering
 \begin{subfigure}[c]{0.49\textwidth}
 \centering
 \includegraphics[width=\textwidth]{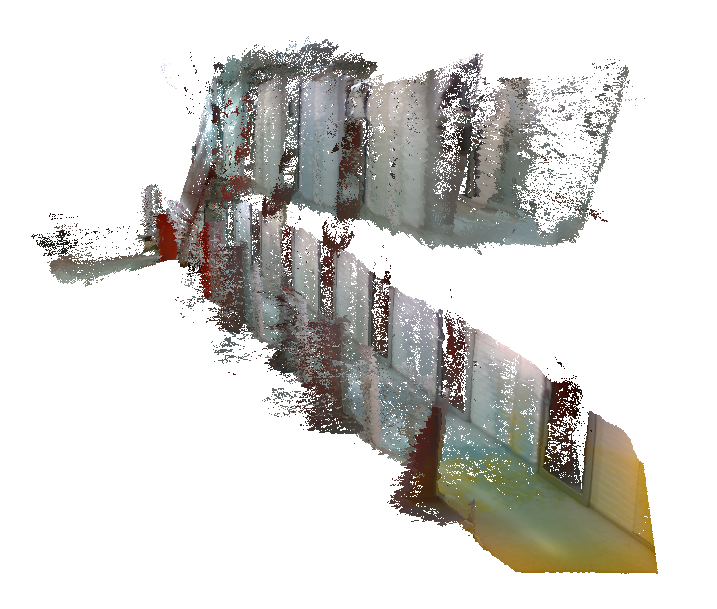}
 \end{subfigure}
 \begin{subfigure}[c]{0.49\textwidth}
 \centering
 \caption{\textit{Lab room}: handheld.}
 \end{subfigure}
 \centering
 \begin{subfigure}[c]{0.49\textwidth}
 \centering
 \caption{\textit{Lab corridor}: handheld.}
 \end{subfigure}
 \caption{Mesh reconstruction results using GCN-SLAM as input to TSDF volume integration from Open3D. Loop closure detection in GCN-SLAM was disabled to demonstrate the accuracy in the tracking alone.}
 \label{fig:mesh}
 \end{figure*}

\jiexiong{Since there is no ground truth for the datasets we collected, these results can only be interpreted qualitatively, and serve as an addition to the quantitative results presented in \cref{sec:quantitative}.} These datasets were selected to show that our method handles difficult scenarios, is robust, and can be used for real-time positioning of a drone. \Cref{fig:trajs} shows the estimated trajectories of GCN-SLAM using ORB versus GCNv2 as keypoints. Note that both methods are evaluated in exactly the same tracking pipeline for fair comparison, i.e. GCNv2 or ORB features is the \textit{only} differences. Refer to the source code for exact details.
\jiexiong{In \cref{fig:trajs_cor}, ORB features are unable to cope with the 180 degree turn at the top right of the trajectory, and in \cref{fig:trajs_park} tracking fails almost immediately.} Futhermore, \cref{fig:trajs_atrium,fig:trajs_kitchen} show that using GCN-SLAM as a basis for drone control improves performance. Specifically, in \cref{fig:trajs_atrium} the position is estimated using only an optical flow sensor whereas in \cref{fig:trajs_kitchen}, GCN-SLAM was used as the positioning source. It is clear that the drone is able to hold its position better, and there is less noise in the latter trajectory. In all four datasets, tracking is maintained with GCNv2, but lost with ORB. We used a remote control to send setpoints to the flight control unit on the drone for control, using the built-in position holding mode.

In \cref{fig:inliers} we further compare the performance of our keypoint extractor to the baseline ORB keypoint extractor. We plot the number of inliers during tracking of the local map for our adapted SLAM system, first with ORB keypoints, and then with GCNv2 keypoints. As the figure illustrates, while there are many more ORB features, our method has a higher percentage of inliers. In addition, as shown in \cref{fig:digest}, GCNv2 results in better distributed features compared with ORB.


\section{Conclusions}
 
In our previous work~\cite{GCN}, we showed that GCN achieves better performance in visual tracking than existing deep learning and classical methods. However, GCN cannot be directly deployed into a real-time SLAM system in an efficient way due to its computational demands and its use of multiple image frames. In this paper, we addressed these issues by proposing a smaller, more efficient version of GCN, called GCNv2, that is readily adaptable to existing SLAM systems. We showed that GCNv2 can be effectively used in a modern feature-based SLAM system to achieve state-of-the-art tracking performance. The robustness and performance of the method was verified by incorporating GCNv2 into GCN-SLAM and using it on-board for positioning on our drone.

\textbf{Limitations}
GCNv2 is trained to predict projective geometry, and not generic feature matching. This is an intentional limitation of scope on our part.
As always with learning-based methods, generalization is an important factor. GCNv2 works relatively well for outdoor scenes, as demonstrated in our experiments (see \cref{fig:trajs_park}) even though the training dataset contains no outdoor data, performance can likely be improved in that environment. Our target here is an indoor setting and we did not investigate this further.

\textbf{Future work}
In the future, we are interested in exploiting semantic information to reject outliers using higher-level information and fusing this information into the motion estimation to improve the capability of our system, especially in environments with non-static objects. We would also like to investigate training GCN in a self-supervised or unsupervised manner, to allow our system to self-improve online, and over time.

    \balance
    \bibliographystyle{IEEEtran}
	\bibliography{mybib}

\end{document}